\documentclass[10pt,twocolumn,letterpaper]{article}
\usepackage{cvpr}
\usepackage{times}
\usepackage{epsfig}
\usepackage{graphicx}
\usepackage{amsmath}
\usepackage{amssymb}

\usepackage{multirow}

\usepackage[breaklinks=true,bookmarks=false]{hyperref}

\cvprfinalcopy 

\def\cvprPaperID{****} 

\begin{document}

\title{Face Detection with the Faster R-CNN}

\author{
Huaizu Jiang\\
University of Massachusetts Amherst\\
Amherst MA 01003\\
{\tt\small hzjiang@cs.umass.edu}
\and
Erik Learned-Miller\\
University of Massachusetts Amherst\\
Amherst MA 01003\\
{\tt\small elm@cs.umass.edu}
}

\maketitle

\begin{abstract}
   The Faster R-CNN~\cite{ren15faster} has recently demonstrated
   impressive results on various object detection benchmarks. By
   training a Faster R-CNN model on the large scale WIDER face
   dataset~\cite{yang16wider}, we report state-of-the-art results 
   on two widely used face detection benchmarks, FDDB and the recently 
   released IJB-A.
\end{abstract}

\section{Introduction}
Deep convolutional neural networks (CNNs) have dominated many tasks of
computer vision. In object detection, region-based CNN detection
methods are now the main paradigm. It is such a rapidly developing
area that three generations of region-based CNN detection models have
been proposed in the last few years, with increasingly better
performance and faster processing speed.

The latest generation, represented by the Faster R-CNN of Ren, He,
Girshick, and Sun~\cite{ren15faster} demonstrates impressive results
on various object detection benchmarks. It is also the foundational
framework for the winning entry of the COCO detection challenge
2015.\footnote{\href{http://mscoco.org/dataset/\#detections-leaderboard}{http://mscoco.org/dataset/\#detections-leaderboard}}
In this report, we demonstrate state-of-the-art face detection results
using the Faster R-CNN on two popular face detection benchmarks, the
widely used Face Detection Dataset and Benchmark
(FDDB)~\cite{fddbTech}, and the more recent IJB-A
benchmark~\cite{klare15pushing}.  We also compare different
generations of region-based CNN object detection models, and compare
to a variety of other recent high-performing detectors.

\section{Overview of the Faster R-CNN}

Since the dominant success of a deeply trained convolutional network (CNN)~\cite{krizhevsky12imagenet} in image classification on the ImageNet Large Scale Visual Recognition Challenge (ILSVRC) 2012, it was wondered that if the same success can be achieved for object detection. The short answer is yes. 

\subsection{Evolution of Region-based CNNs for Object Detection}
Girshick~\etal~\cite{girshick14rich} introduced a region-based CNN (R-CNN) for object detection. The pipeline consists of two stages. In the first, a set of category-independent object proposals are generated, using selective search~\cite{uijlings13selective}. In the second refinement stage, the image region within each proposal is warped to a fixed size (\eg, $227\times227$ for the AlexNet~\cite{krizhevsky12imagenet}) and then mapped to a 4096-dimensional feature vector, which is fed into a classifier and also into a regressor that refines the position of the detection. 

The significance of the R-CNN is that it brings the high accuracy of
CNNs on classification tasks to the problem of object detection.  Its
success is largely due to transferring the supervised pre-trained
image representation for image classification to object detection.

The R-CNN, however, requires a forward pass through the convolutional
network for \emph{each} object proposal in order to extract features, leading
to a heavy computational burden. To mitigate this problem, two approaches,
the SPPnet~\cite{he14spatial} and the Fast R-CNN~\cite{girshick15fast} have
been 
proposed. Instead of feeding each warped proposal image region to
the CNN, the SPPnet and the Fast R-CNN run through the CNN exactly \emph{once}
for the entire input image. After projecting the proposals to
convolutional feature maps, a fixed length feature vector can be
extracted for each proposal in a manner similar to spatial pyramid
pooling. The Fast R-CNN is a special case of the SPPnet, which uses a single
spatial pyramid pooling layer, \ie, the region of interest (RoI) pooling layer,
and thus allows end-to-end fine-tuning of a pre-trained ImageNet
model. This is the key to its better performance relative to the 
 original R-CNN.

 \begin{table*}[t]
   \centering
   \caption{Comparisons of the \emph{entire} pipeline of different
     region-based object detection methods. (Both Faceness~\cite{yang15from} and DeepBox~\cite{kuo15deepbox}
     rely on the output of EdgeBox. Therefore their entire running
     time should include the processing time of EdgeBox.)}
   \begin{tabular}{c|c|c|c|c}
   \hline
   \multicolumn{2}{c|}{~} & {~~~~R-CNN~~~~} & Fast R-CNN & Faster R-CNN \\
   \hline
   \multirow{3}{*}{proposal stage} & \multirow{3}{*}{time} & \multicolumn{2}{c|}{EdgeBox: 2.73s} & \multirow{4}{*}{0.32s}\\
   & & \multicolumn{2}{c|}{Faceness: 9.91s (+ 2.73s = 12.64s)} & \\
   & & \multicolumn{2}{c|}{DeepBox: 0.27s (+ 2.73s = 3.00s)} & \\
   \hline
   \multirow{3}{*}{refinement stage} & input to CNN & cropped proposal image & input image \& proposals & input image \\
   \cline{2-5}
   & \#forward thru. CNN & \#proposals & 1 & 1 \\
   \cline{2-5}
   & time & 7.04s & 0.21s & 0.06s\\
   \hline
   \multirow{3}{*}{total} & \multirow{3}{*}{time} & R-CNN + EdgeBox: 9.77s & Fast R-CNN + EdgeBox: 2.94s & \multirow{3}{*}{0.38s} \\
    & & R-CNN + Faceness: 19.68s & Fast R-CNN + Faceness: 12.85s & \\
    & & R-CNN + DeepBox: 10.04s & Fast R-CNN + DeepBox: 3.21s & \\
   \hline
   \end{tabular}
   \label{tab:rcnnComp}
\end{table*}

Both the R-CNN and the Fast R-CNN (and the SPPNet) rely on the input
generic object proposals, which usually come from a hand-crafted model
such as selective search~\cite{uijlings13selective},
EdgeBox~\cite{dollar15fast}, etc. There are two main issues with this
approach. The first, as shown in image classification and object
detection, is that (deeply) learned representations often generalize
better than hand-crafted ones. The second is that the computational
burden of proposal generation dominate the processing time of the
entire pipeline (\eg, 2.73 seconds
for EdgeBox in our experiments). Although there are now deeply trained models for
proposal generation, \eg DeepBox~\cite{kuo15deepbox} (based on the
Fast R-CNN framework), its processing time is still not negligible.

To reduce the computational burden of proposal generation, the Faster
R-CNN was proposed. It consists of two modules. The first, called the
Regional Proposal Network (RPN), is a fully convolutional network for
generating object proposals that will be fed into the second
module. The second module is the Fast R-CNN detector whose purpose is
to refine the proposals. The key idea is to \emph{share} the same
convolutional layers for the RPN and Fast R-CNN detector up to their
own fully connected layers. Now the image only passes through the CNN
once to produce and then refine object proposals. More importantly,
thanks to the sharing of convolutional layers, it is possible to use a
very deep network (\eg, VGG16~\cite{simonyan14very}) to generate
high-quality object proposals.

The key differences of the R-CNN, the Fast R-CNN, and the Faster R-CNN
are summarized in Table~\ref{tab:rcnnComp}. The running time of
different modules are reported on the FDDB dataset~\cite{fddbTech},
where the typical resolution of an image is about $350\times450$. The
code was run on a server equipped with an Intel Xeon CPU E5-2697 of
2.60GHz and an NVIDIA Tesla K40c GPU with 12GB memory. We can clearly
see that the entire running time of the Faster R-CNN is significantly
lower than for both the R-CNN and the Fast R-CNN.

\subsection{The Faster R-CNN}
In this section, we briefy introduce the key aspects of the Faster
R-CNN. We refer readers to the original paper~\cite{ren15faster} for
more technical details.

In the RPN, the convolution layers of a pre-trained network are followed by
a $3\times3$ convolutional layer. This corresponds to mapping a large
spatial window or {\em receptive field} (\eg, $228\times228$ for VGG16)
in the input image to a low-dimensional feature
vector at a center stride (\eg, 16 for VGG16). Two $1\times1$
convolutional layers are then added for \emph{classification} and
\emph{regression} branches of all spatial windows.

To deal with different scales and aspect ratios of objects,
\emph{anchors} are introduced in the RPN. An anchor is at each sliding
location of the convolutional maps and thus at the center of each
spatial window. Each anchor is associated with a scale and an aspect
ratio. Following the default setting of~\cite{ren15faster}, we use 3
scales ($128^2, 256^2$, and $512^2$ pixels) and 3 aspect ratios ($1:1,
1:2$, and $2:1$), leading to $k=9$ anchors at each location. Each
proposal is parameterized relative to an anchor. Therefore, for a
convolutional feature map of size $W\times H$, we have at most $WHk$
possible proposals. We note that the same features of each sliding
location are used to regress $k=9$ proposals, instead of extracting
$k$ sets of features and training a single regressor.
                                                                                                                                                                                                                                                                                                                                                                                                                                                                                                                                                                                                                                                                                                                                                                                                            t
Training of the RPN can be done in an end-to-end manner using
stochastic gradient descent (SGD) for both classification and
regression branches. For the entire system, we have to take care of
both the RPN and Fast R-CNN modules since they share convolutional
layers. In this paper, we adopt the approximate joint learning
strategy proposed in~\cite{ren16faster}. The RPN and Fast R-CNN are
trained end-to-end as they are independent. Note that the input of the
Fast R-CNN is actually dependent on the output of the RPN. For the
exact joint training, the SGD solver should also consider the
derivates of the RoI pooling layer in the Fast R-CNN with respect to
the coordinates of the proposals predicted by the RPN. However, as
pointed out by~\cite{ren16faster}, it is not a trivial optimization
problem.

\begin{figure}[t]
\centering
\renewcommand{\arraystretch}{0.6}
\renewcommand{\tabcolsep}{.05mm}
\begin{tabular}{cc}
   \includegraphics[height=0.36\linewidth,keepaspectratio]{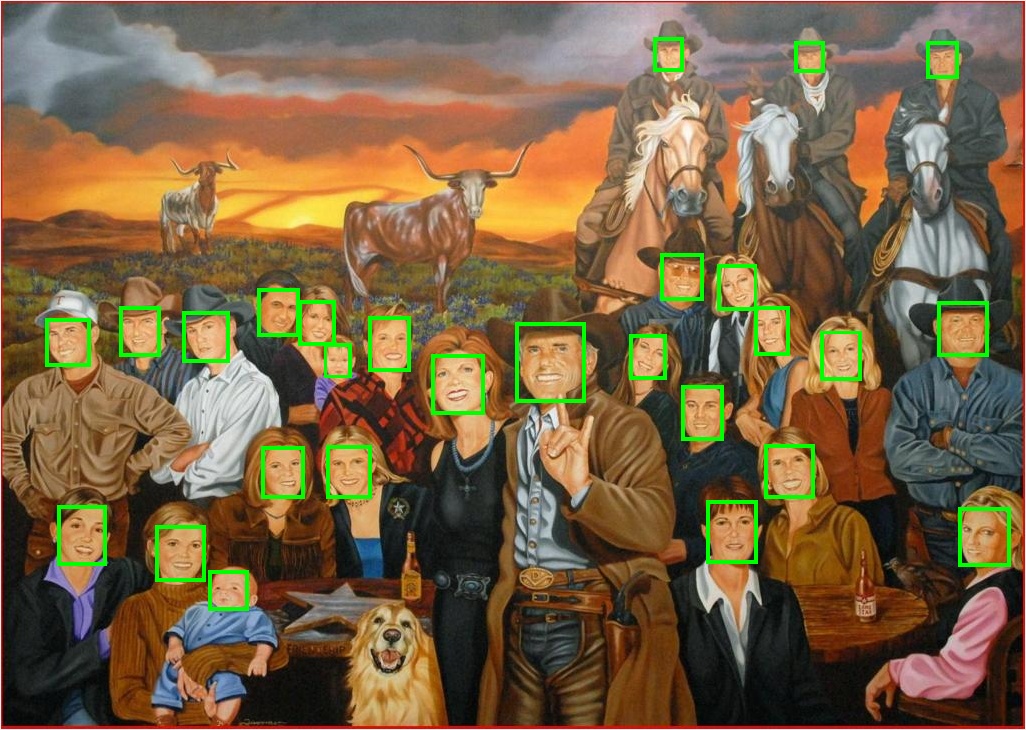} &
   \includegraphics[height=0.36\linewidth,keepaspectratio]{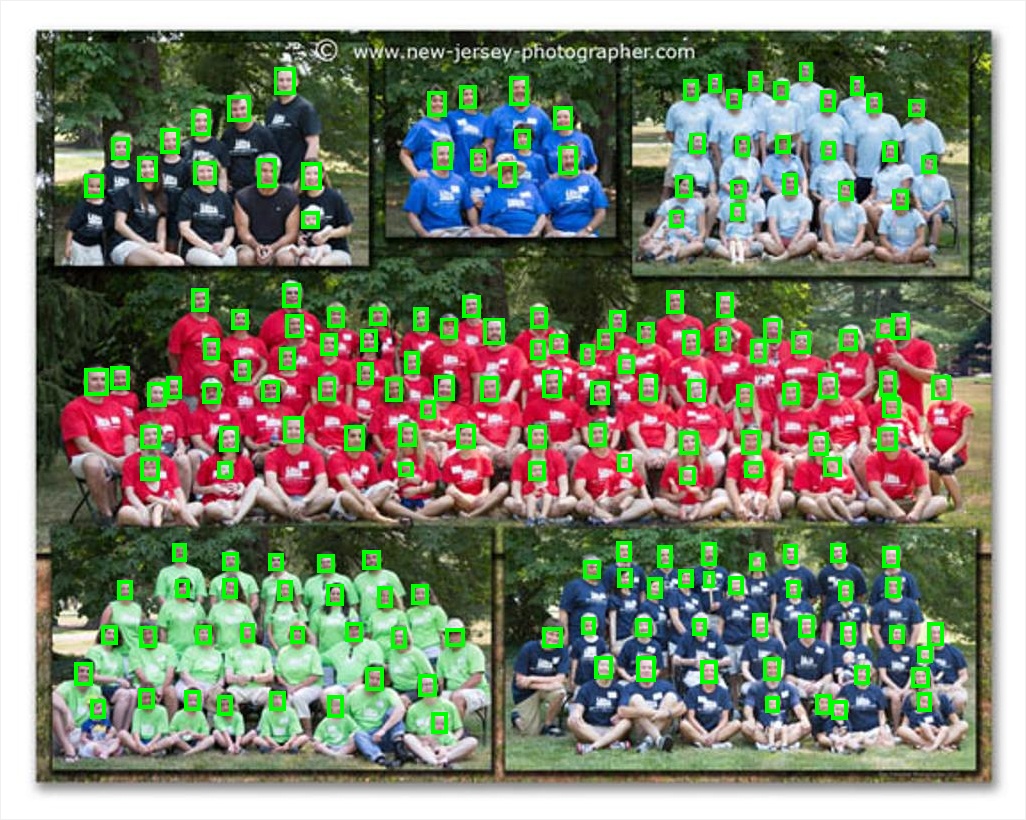} \\
   \includegraphics[height=0.36\linewidth,keepaspectratio]{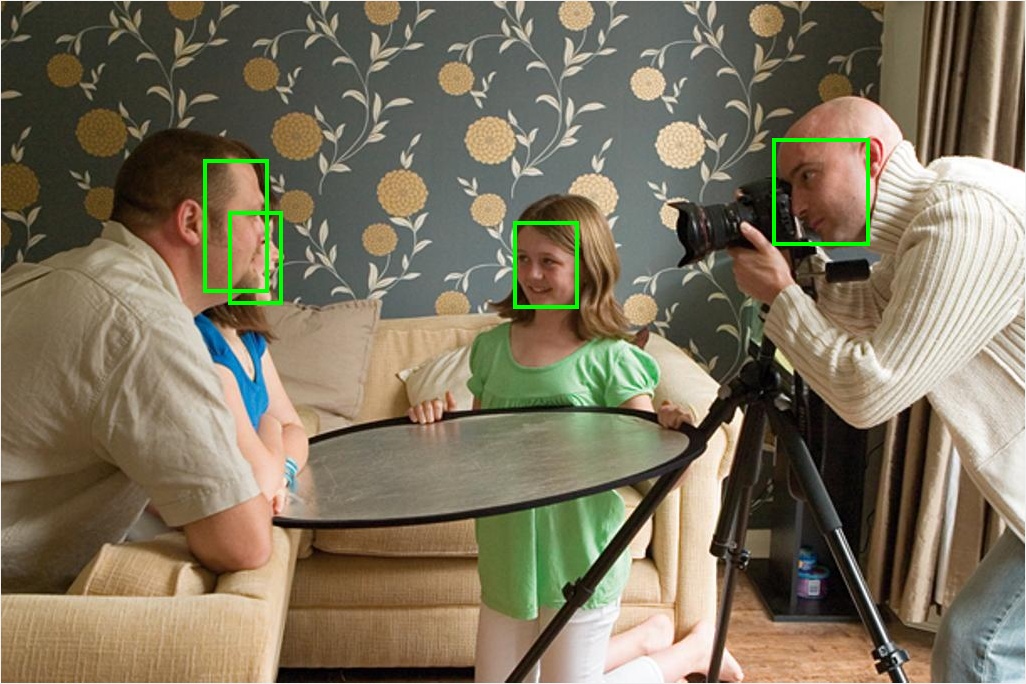} &
   \includegraphics[height=0.36\linewidth,keepaspectratio]{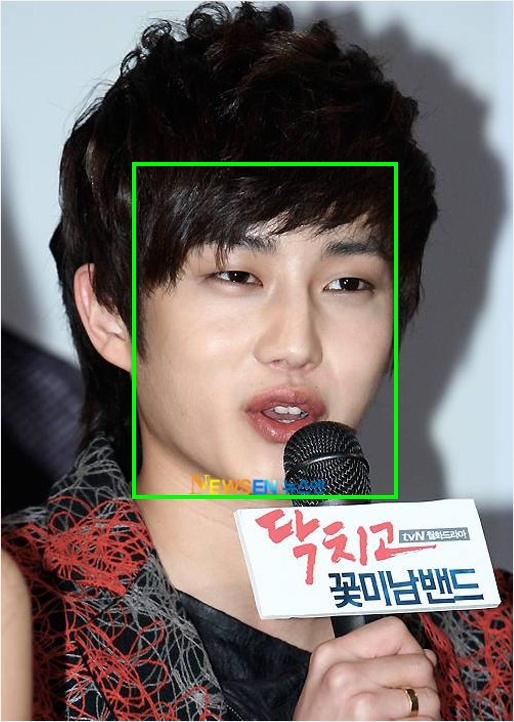} \\
   \includegraphics[height=0.36\linewidth,keepaspectratio]{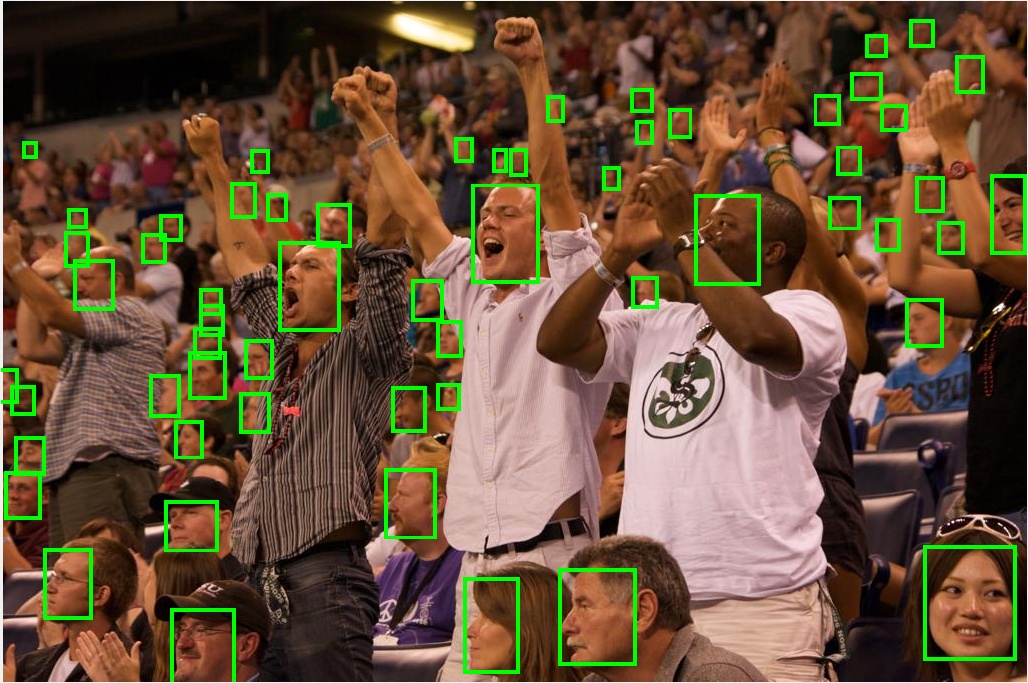} &
   \includegraphics[height=0.36\linewidth,keepaspectratio]{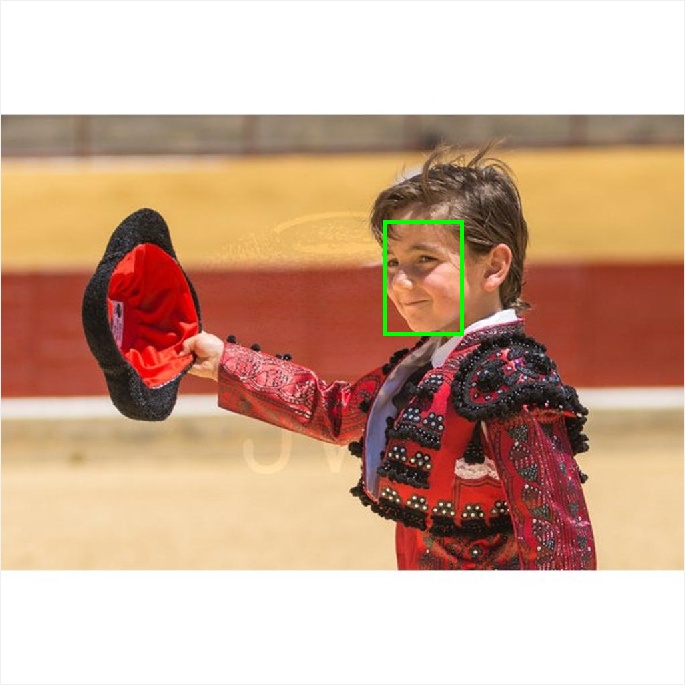} \\
\end{tabular}
\caption{Sample images in the WIDER face dataset, where green bounding boxes are ground-truth annotations.}
\label{fig:sampleWider}
\end{figure}

\section{Experiments}
In this section, we report experiments on comparisons of region
proposals and also on end-to-end performance of top face detectors.
\subsection{Setup}
We train a Faster R-CNN face detection model on the recently released
WIDER face dataset~\cite{yang16wider}. There are 12,880 images and
159,424 faces in the training set. In Fig.~\ref{fig:sampleWider}, we
demonstrate some randomly sampled images of the WIDER dataset. We can see
that there exist great variations in scale, pose, and the number of
faces in each image, making this dataset challenging.

We train the face detection model based on a pre-trained ImageNet model, VGG16~\cite{simonyan14very}. We randomly sample one image per batch for training. In order to fit it in the GPU memory, it is resized based on the ratio $1024/\max(w, h)$, where $w$ and $h$ are the width and height of the image, respectively. We run the SGD solver 50k iterations with a base learning rate of 0.001 and run another 20K iterations reducing the base learning rate to 0.0001.\footnote{We use the author released Python implementation \href{https://github.com/rbgirshick/py-faster-rcnn}{https://github.com/rbgirshick/py-faster-rcnn}.}

\begin{figure}[t]
   \centering
   \renewcommand{\arraystretch}{0.6}
   \renewcommand{\tabcolsep}{.2mm}
   \begin{tabular}{cccc}
      \includegraphics[width=0.5\linewidth,keepaspectratio]{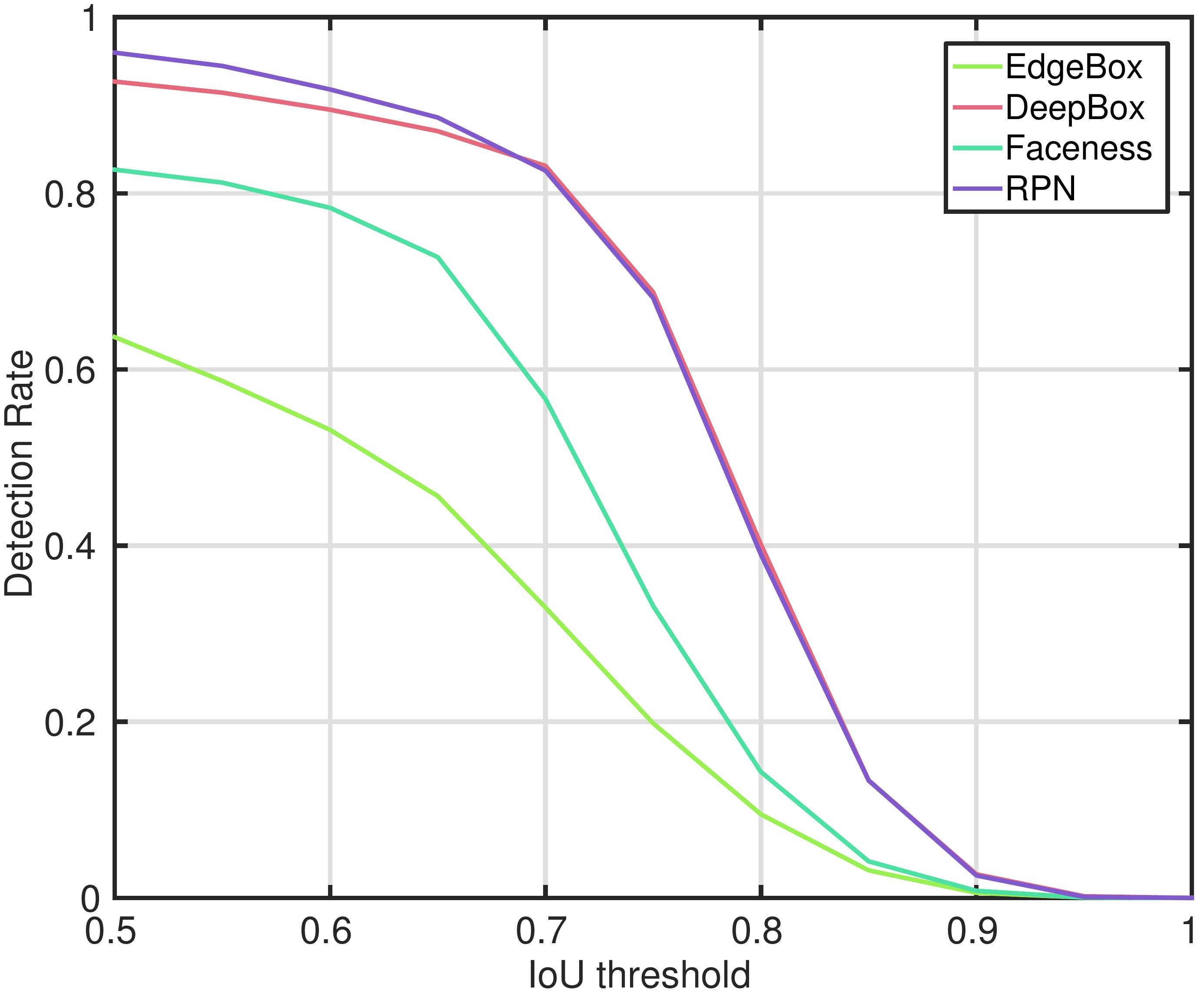} &
      \includegraphics[width=0.5\linewidth,keepaspectratio]{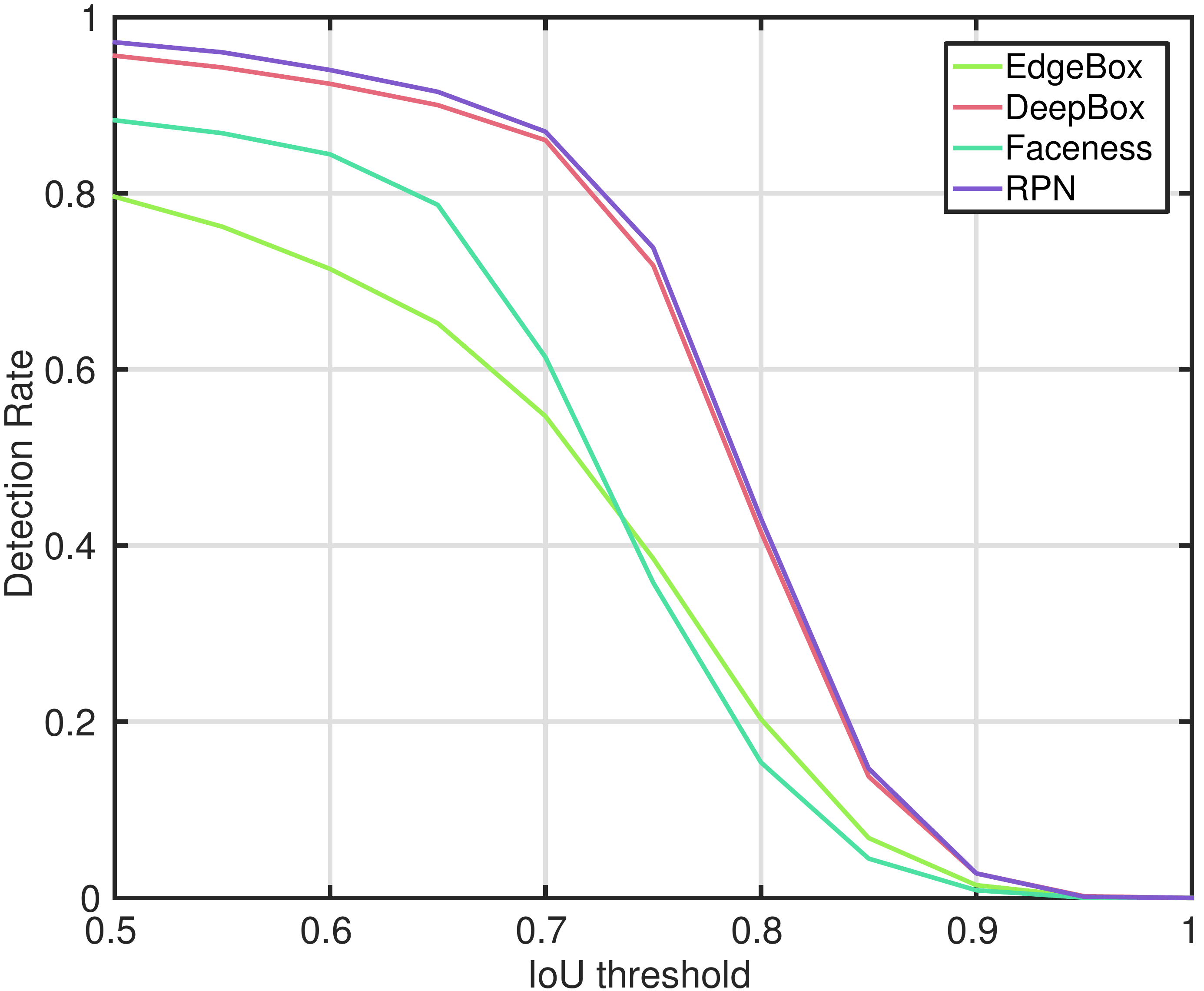} \\
      (a) 100 Proposals & (b) 300 Proposals \\
      \includegraphics[width=0.5\linewidth,keepaspectratio]{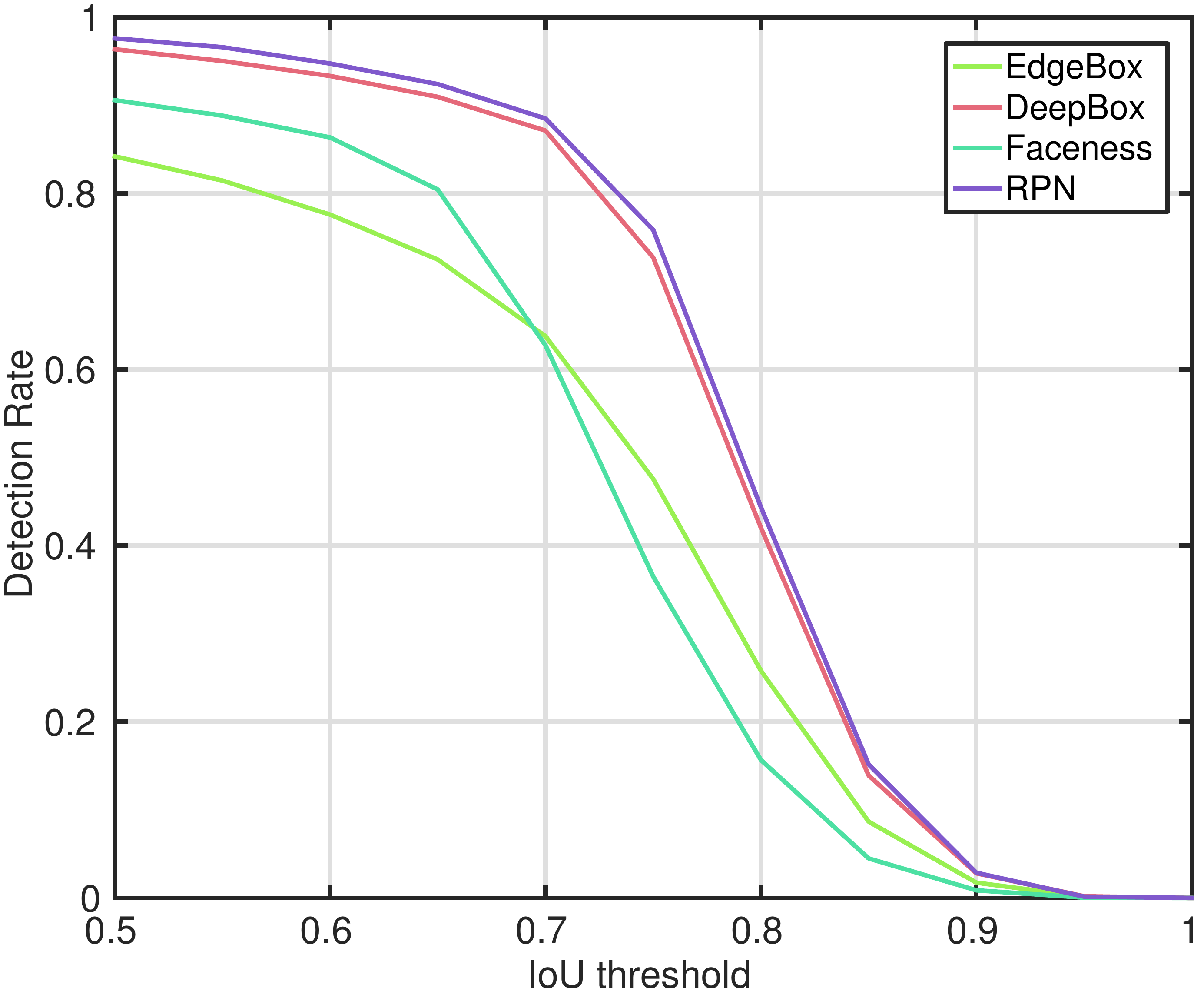} &
      \includegraphics[width=0.5\linewidth,keepaspectratio]{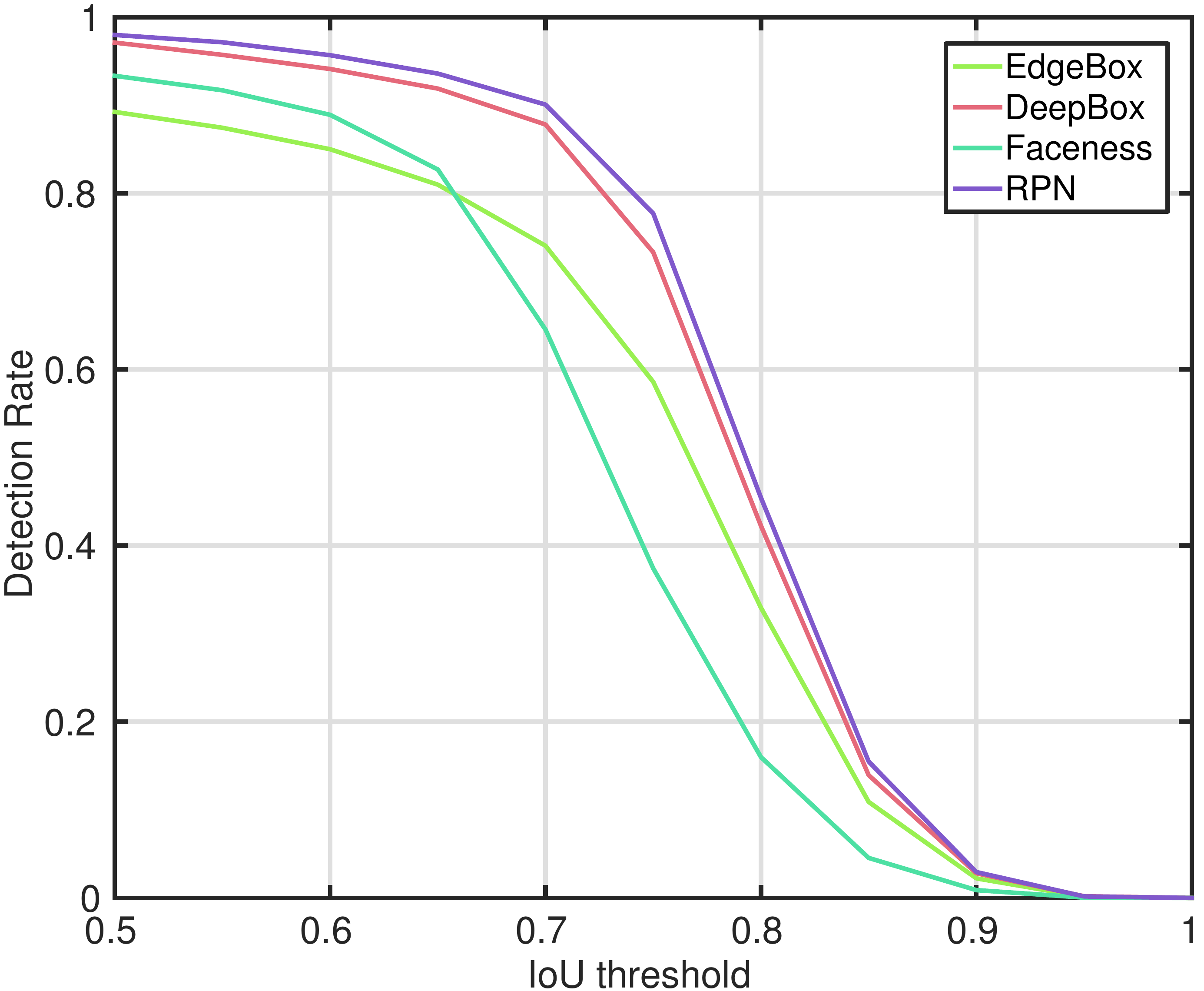} \\
      (c) 500 proposals & (d) 1000 proposals
   \end{tabular}
   \caption{Comparisons of face proposals on FDDB using different methods.}
   \label{fig:propCompFddb}
\end{figure}

We test the trained face detection model on two benchmark datasets, FDDB~\cite{fddbTech} and IJB-A~\cite{klare15pushing}. There are 10 splits in both FDDB and IJB-A. For testing, we resize the input image based on the ratio $\min(600/\min(w, h), 1024/\max(w, h))$. For the RPN, we use only the 
top 300 face proposals to balance efficiency and accuracy.

For FDDB, we directly test the model trained on WIDER. For IJB-A, it
is necessary to fine-tune the face detection model due to the different
annotation styles between WIDER and IJB-A.
In WIDER, the face annotations are specified tightly around the facial
region while annotations in IJB-A include larger areas (\eg, hair). We
fine-tune the face detection model on the training images of each
split of IJB-A using only 10,000 iterations. In the first 5,000 iterations,
the base learning rate is 0.001 and it is reduced to 0.0001 in the
last 5,000.  Note that there are more then 15,000 training images in each
split. We run only 10,000 iterations of fine-tuning to adapt the
regression branches of the Faster R-CNN model trained on WIDER to the
annotation styles of IJB-A.

There are two criteria for quantatitive comparison for the FDDB
benchmark.  For the discrete scores, each detection is considered to
be positive if its intersection-over-union (IoU) ratioe with its
one-one matched ground-truth annotation is greater than 0.5. By
varying the threshold of detection scores, we can generate a set of
true positives and false positives and report the ROC curve. For the
more restrictive continuous scores, the true positives are weighted by
the IoU scores. On IJB-A, we use the discrete score setting and
report the true positive rate based on the normalized false positive rate
per image instead of the total number of false positives.

\begin{figure}
   \centering
   \includegraphics[width=0.95\linewidth,keepaspectratio]{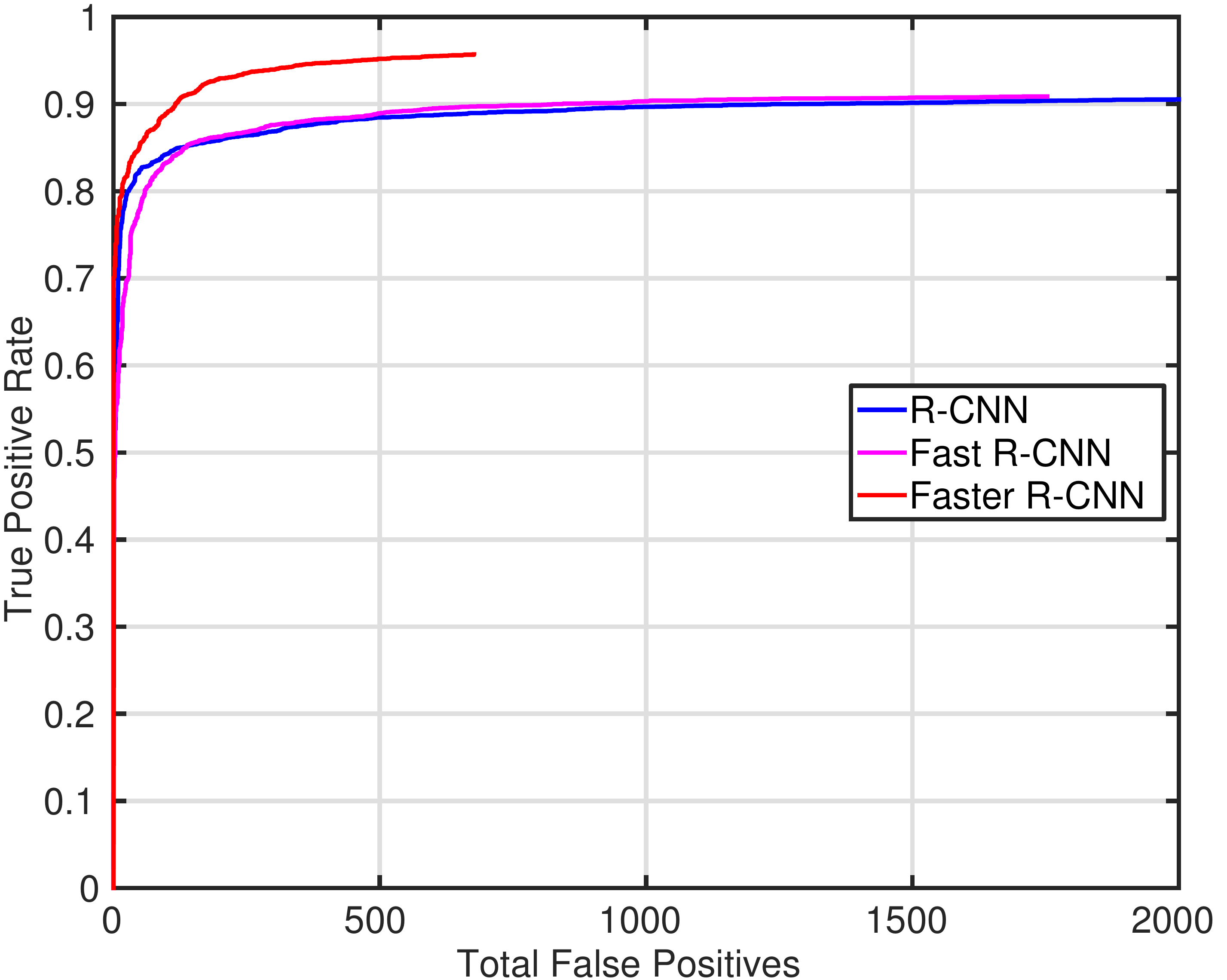} \\
   \caption{Comparisons of region-based CNN object detection methods for face detection on FDDB.}
   \label{fig:rcnnCompFddb}
\end{figure}

\subsection{Comparison of Face Proposals}
We compare the RPN with other approaches including
EdgeBox~\cite{dollar15fast}, Faceness~\cite{yang15from}, and
DeepBox~\cite{kuo15deepbox} on FDDB. EdgeBox evaluates the objectness
score of each proposal based on the distribution of edge responses
within it in a sliding window fashion. Both Faceness and DeepBox
re-rank other object proposals, \eg, EdgeBox. In Faceness, five CNNs
are trained based on attribute annotations of facial parts including
hair, eyes, nose, mouth, and beard. The Faceness score of each proposal is
then computed based on the response maps of different
networks. DeepBox, which is based on the Fast R-CNN framework,
re-ranks each proposal based on the region-pooled features. We re-train a DeepBox model for face proposals on the WIDER training set.

\begin{figure*}[t]
\centering
\begin{tabular}{cc}
   \includegraphics[width=0.46\linewidth,keepaspectratio]{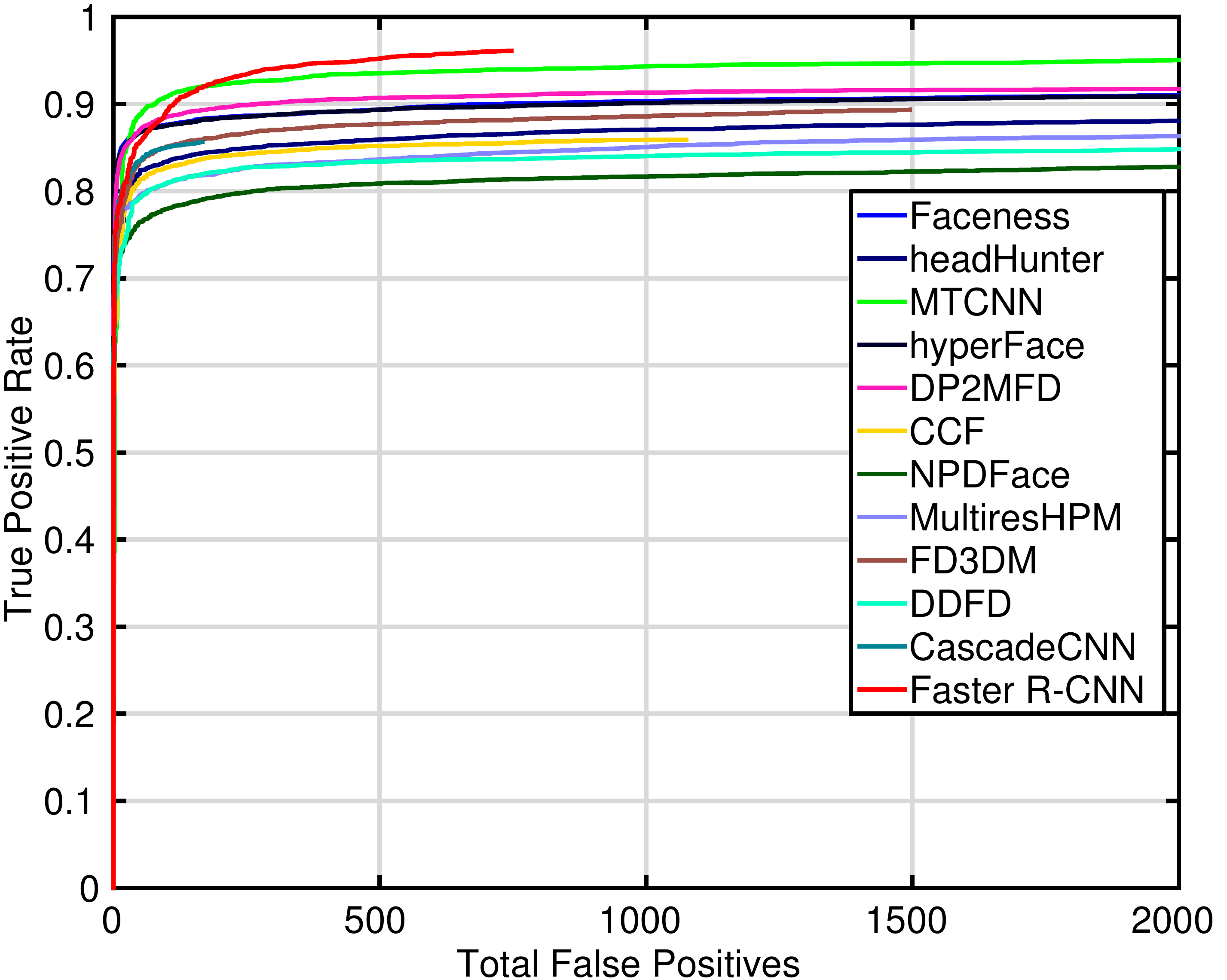} &
   \includegraphics[width=0.46\linewidth,keepaspectratio]{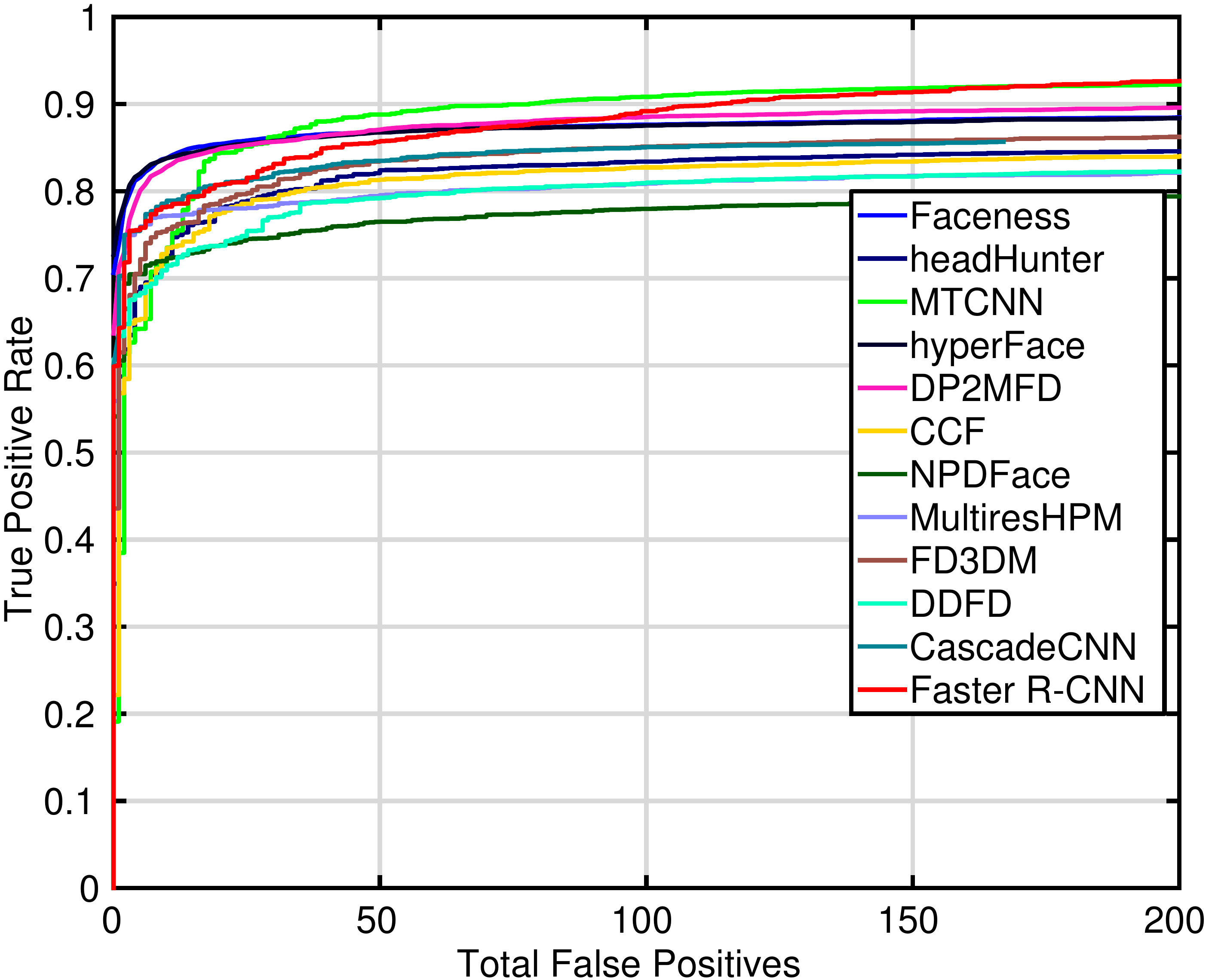} \\
   (a) & (b) \\
   \includegraphics[width=0.46\linewidth,keepaspectratio]{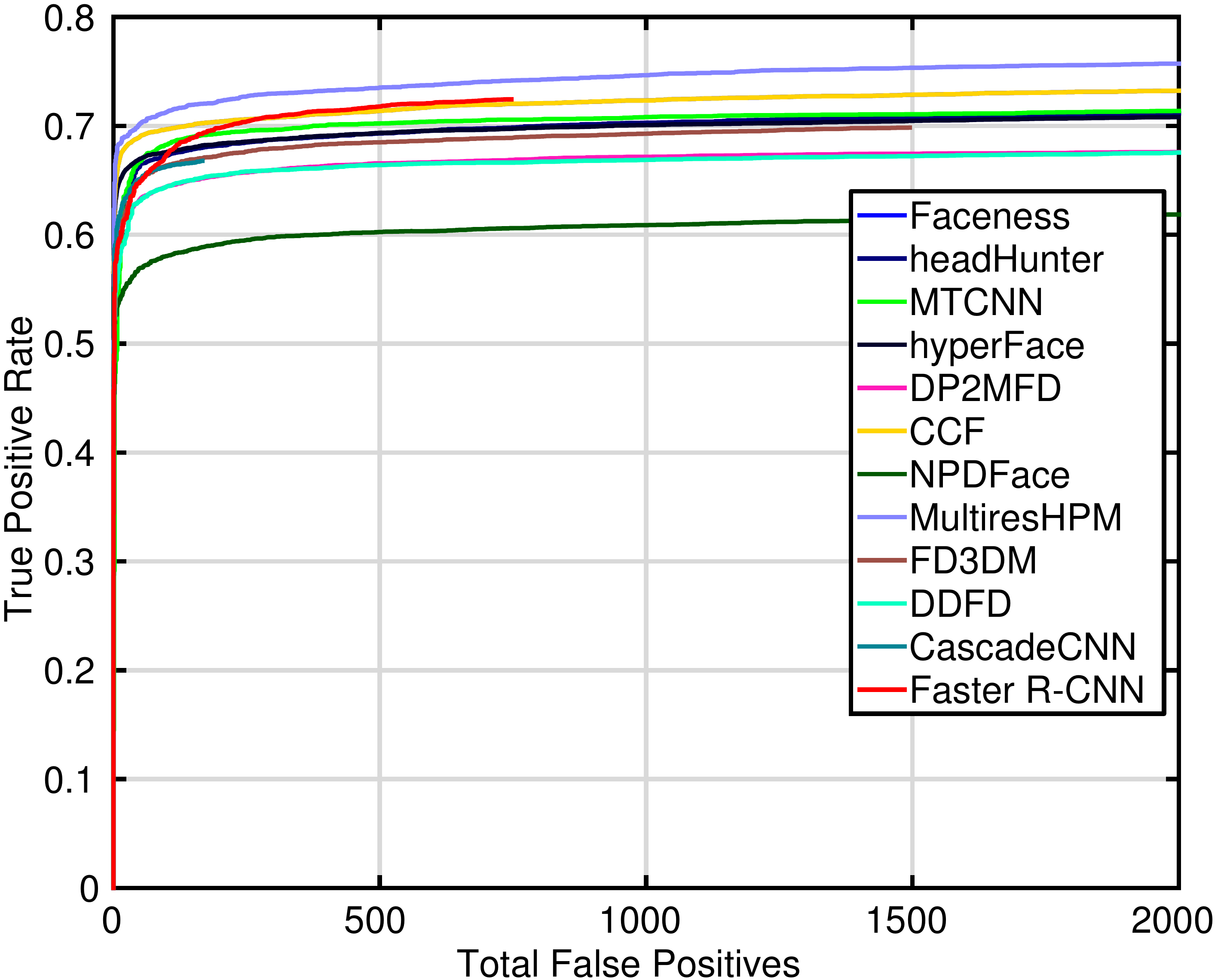} &
   \includegraphics[width=0.46\linewidth,keepaspectratio]{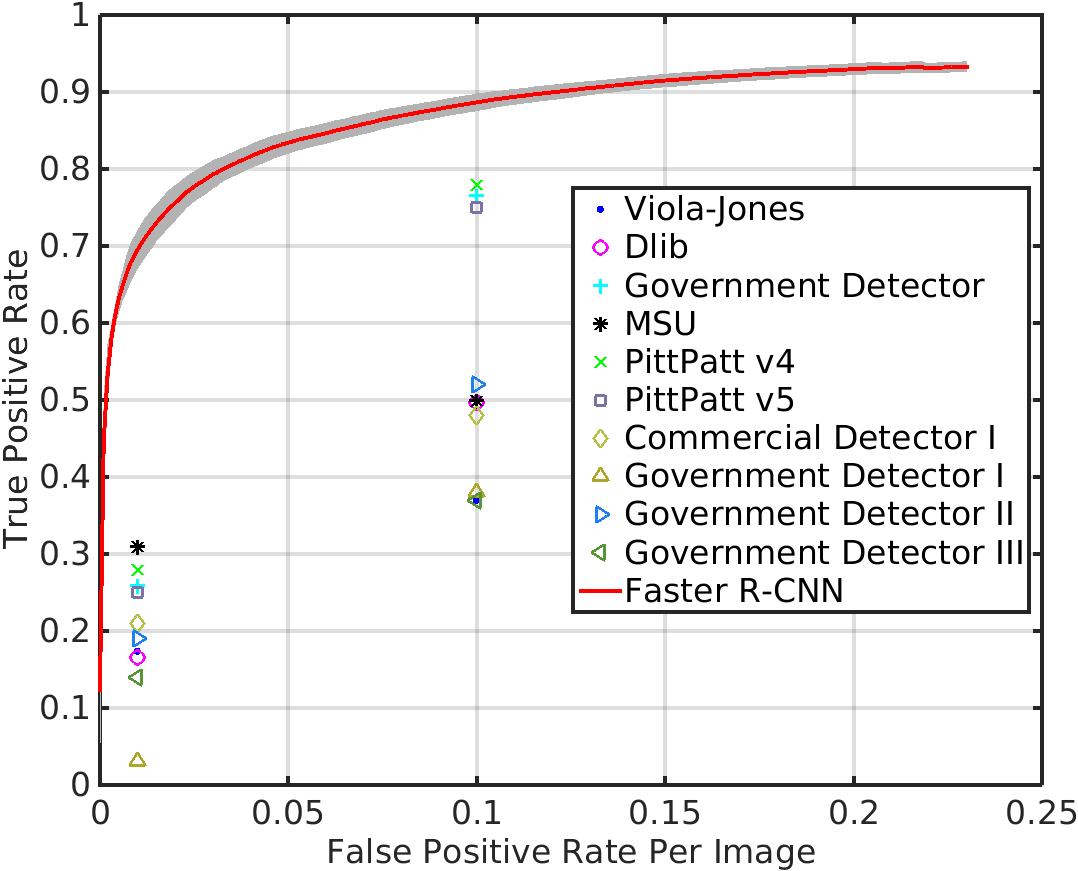} \\
   (c) & (d) \\
\end{tabular}
\caption{Comparisons of face detection with state-of-the-art methods on (a) ROC curves on FDDB with discrete scores, (b) ROC curves on FDDB with discrete scores using less false positives, (c) ROC curves on FDDB with continuous scores, and (d) results on IJB-A dataset.}
\label{fig:compSota}
\end{figure*}

\begin{figure*}
\centering
\renewcommand{\arraystretch}{0.6}
\renewcommand{\tabcolsep}{.05mm}
\begin{tabular}{cccccc}
   \includegraphics[width=0.16\linewidth,keepaspectratio]{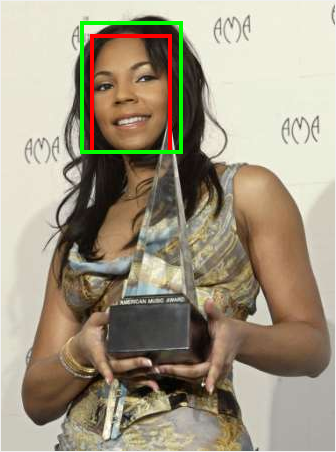} &
\includegraphics[width=0.16\linewidth,keepaspectratio]{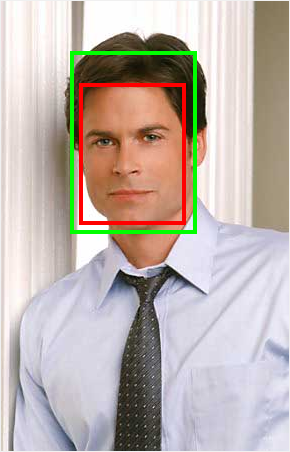} &
\includegraphics[width=0.16\linewidth,keepaspectratio]{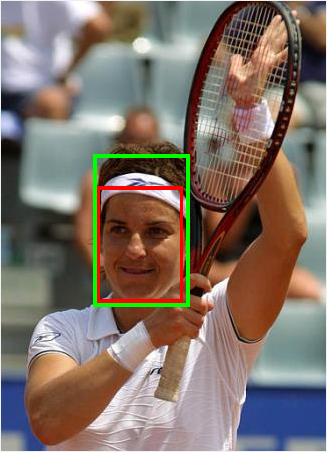} &
\includegraphics[width=0.16\linewidth,keepaspectratio]{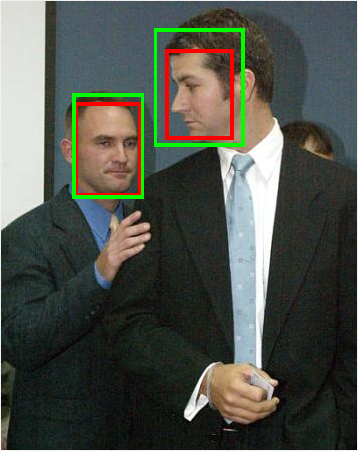} &
\includegraphics[width=0.16\linewidth,keepaspectratio]{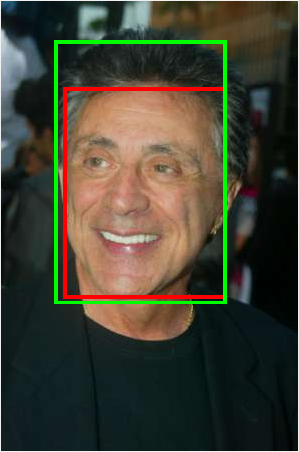} &
\includegraphics[width=0.16\linewidth,keepaspectratio]{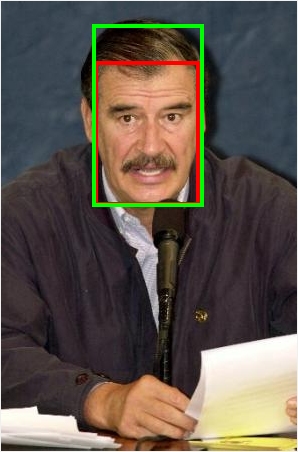} \\
\includegraphics[width=0.16\linewidth,keepaspectratio]{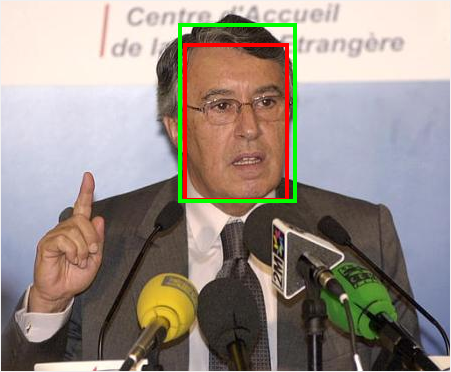} &
\includegraphics[width=0.16\linewidth,keepaspectratio]{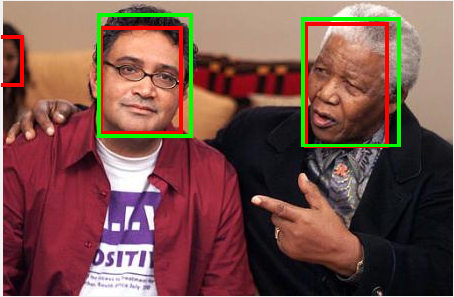} &
\includegraphics[width=0.16\linewidth,keepaspectratio]{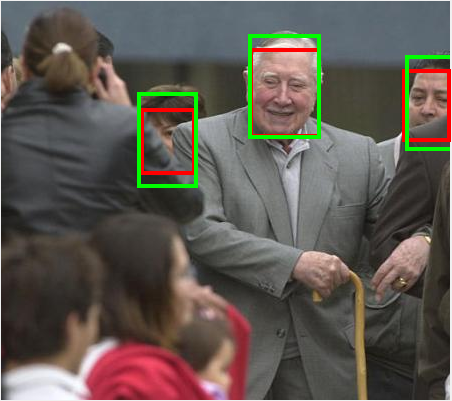} &
\includegraphics[width=0.16\linewidth,keepaspectratio]{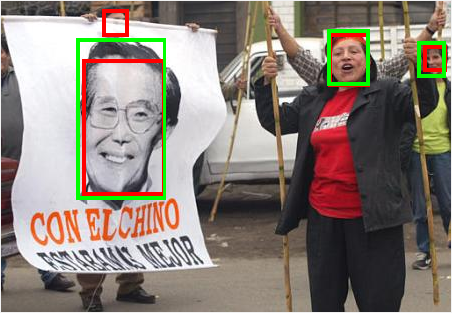} &
\includegraphics[width=0.16\linewidth,keepaspectratio]{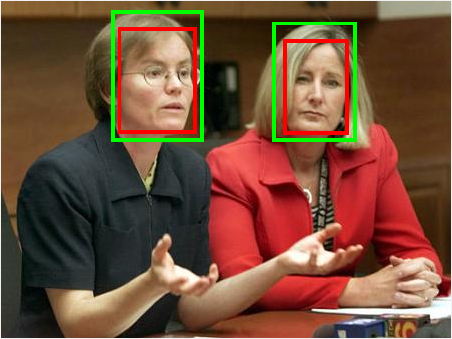} &
\includegraphics[width=0.16\linewidth,keepaspectratio]{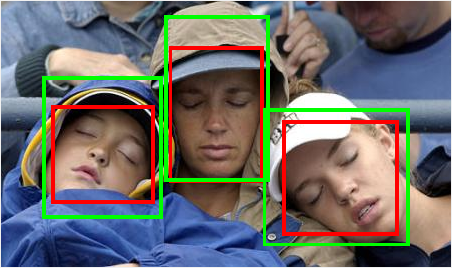} \\
\includegraphics[width=0.16\linewidth,keepaspectratio]{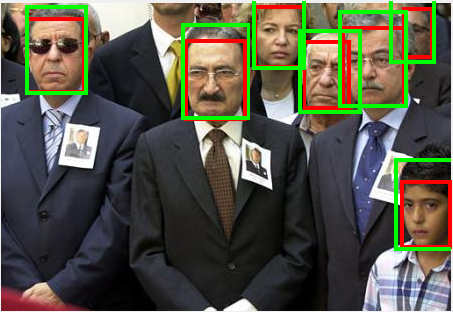} &
\includegraphics[width=0.16\linewidth,keepaspectratio]{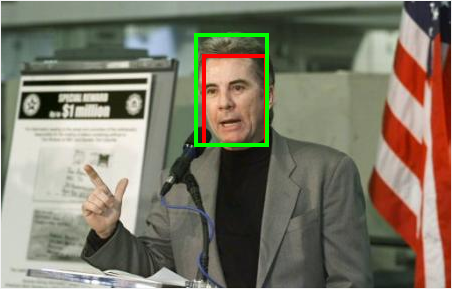} &
\includegraphics[width=0.16\linewidth,keepaspectratio]{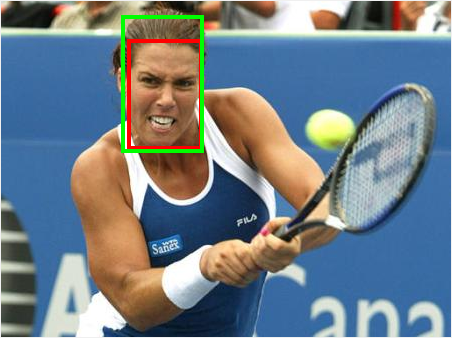} &
\includegraphics[width=0.16\linewidth,keepaspectratio]{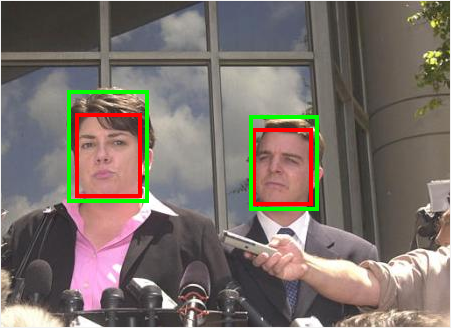} &
\includegraphics[width=0.16\linewidth,keepaspectratio]{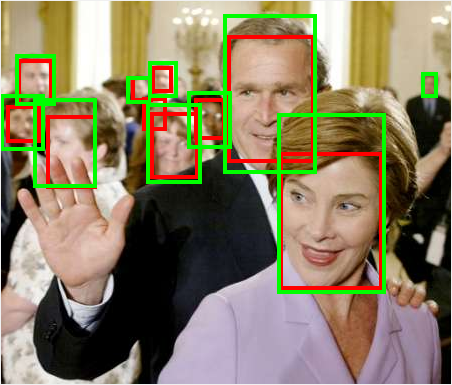} &
\includegraphics[width=0.16\linewidth,keepaspectratio]{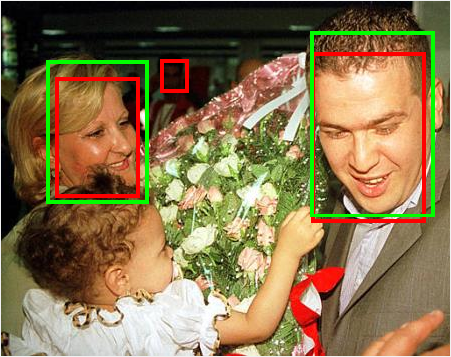} \\
\end{tabular}
\caption{Sample detection results on the FDDB dataset, where {\color{green}green} bounding boxes are ground-truth annotations and {\color{red} red} bounding boxes are detection results of the Faster R-CNN.}
\label{fig:detsFddb}
\end{figure*}

\begin{figure*}
\centering
\renewcommand{\arraystretch}{0.6}
\renewcommand{\tabcolsep}{.05mm}
\begin{tabular}{cccccc}
   \includegraphics[width=0.16\linewidth,keepaspectratio]{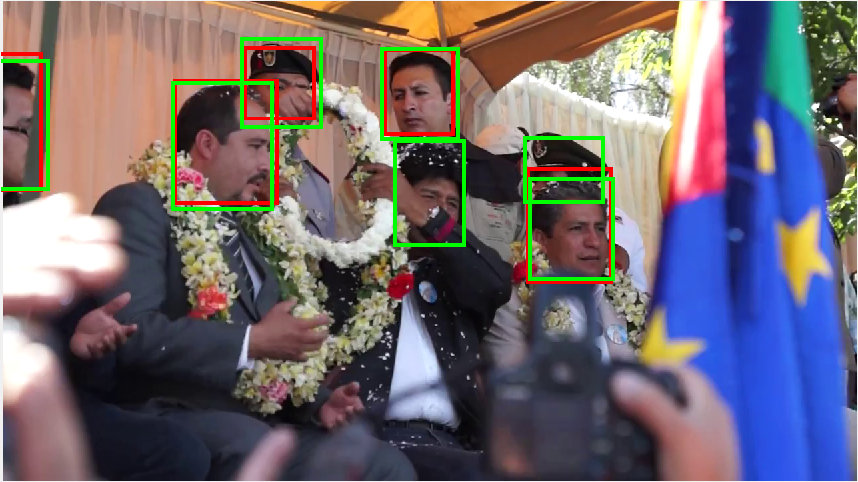} &
   \includegraphics[width=0.16\linewidth,keepaspectratio]{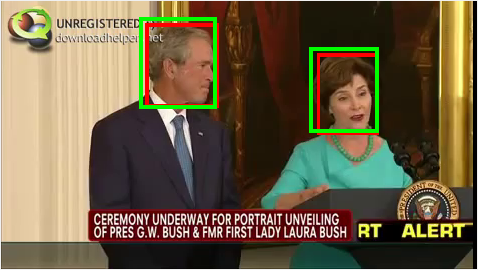} &
   \includegraphics[width=0.16\linewidth,keepaspectratio]{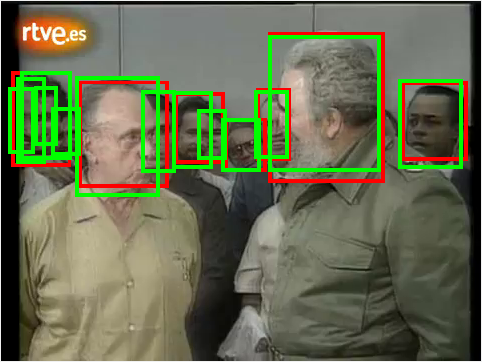} &
   \includegraphics[width=0.16\linewidth,keepaspectratio]{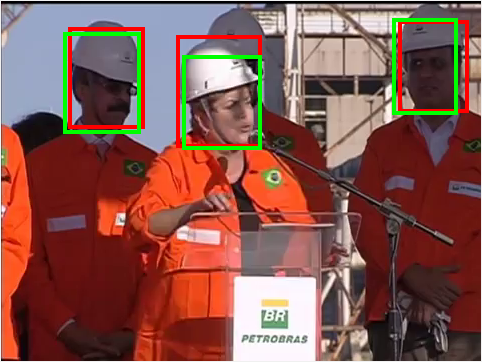} &
   \includegraphics[width=0.16\linewidth,keepaspectratio]{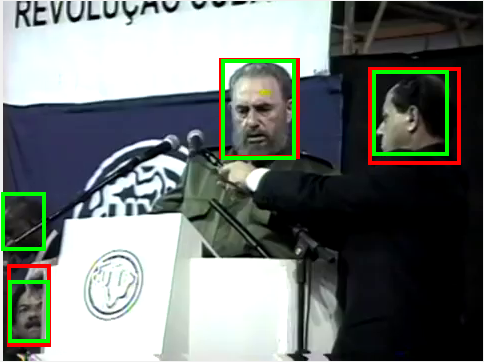} &
   \includegraphics[width=0.16\linewidth,keepaspectratio]{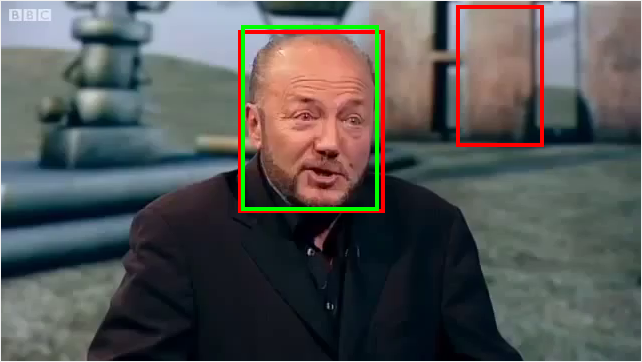} \\
   \includegraphics[width=0.16\linewidth,keepaspectratio]{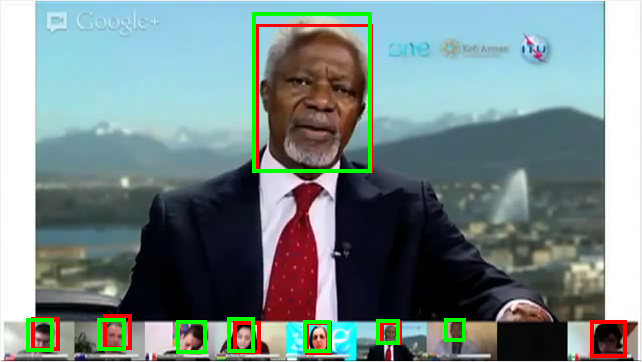} &
   \includegraphics[width=0.16\linewidth,keepaspectratio]{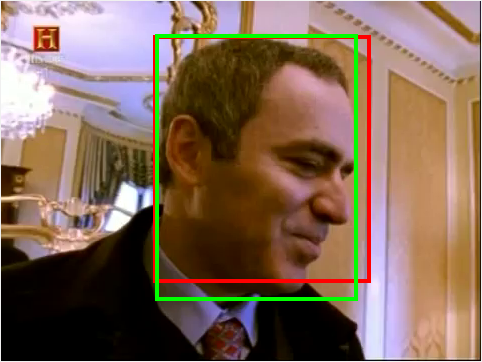} &
   \includegraphics[width=0.16\linewidth,keepaspectratio]{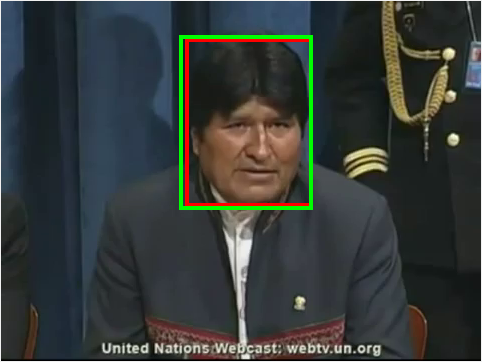} &
   \includegraphics[width=0.16\linewidth,keepaspectratio]{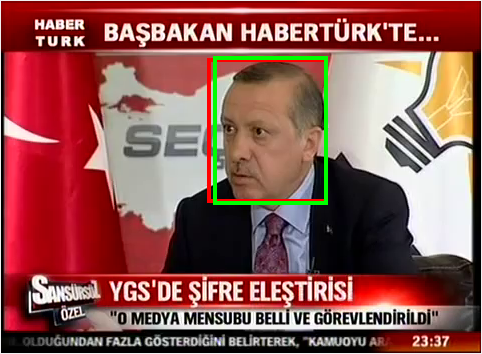} &
   \includegraphics[width=0.16\linewidth,keepaspectratio]{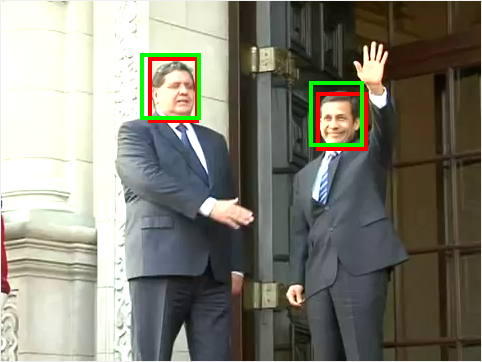} &
   \includegraphics[width=0.16\linewidth,keepaspectratio]{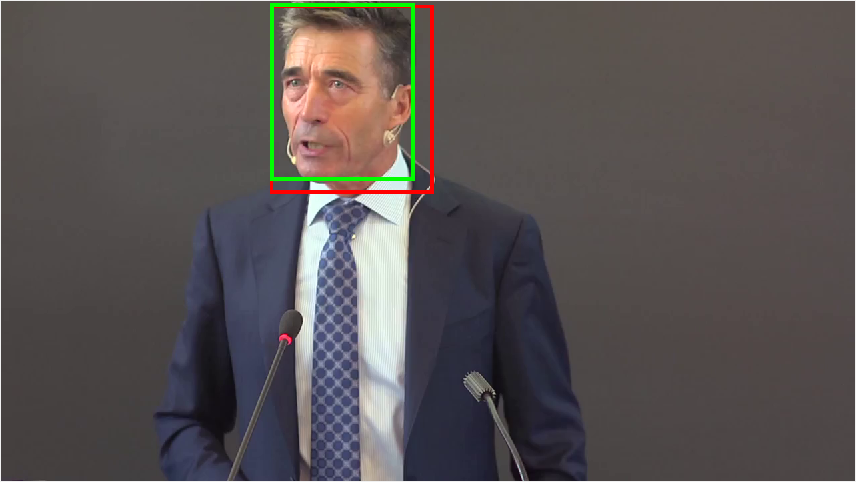} \\
   \includegraphics[width=0.16\linewidth,keepaspectratio]{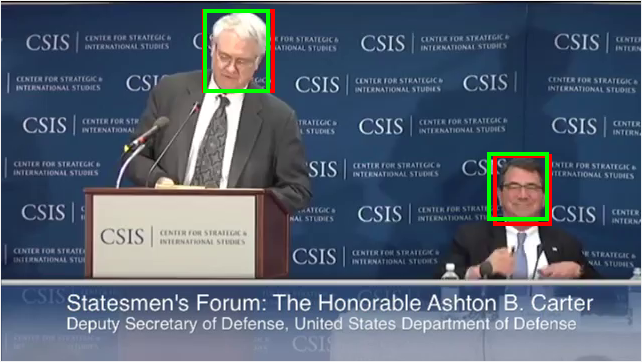} &
   \includegraphics[width=0.16\linewidth,keepaspectratio]{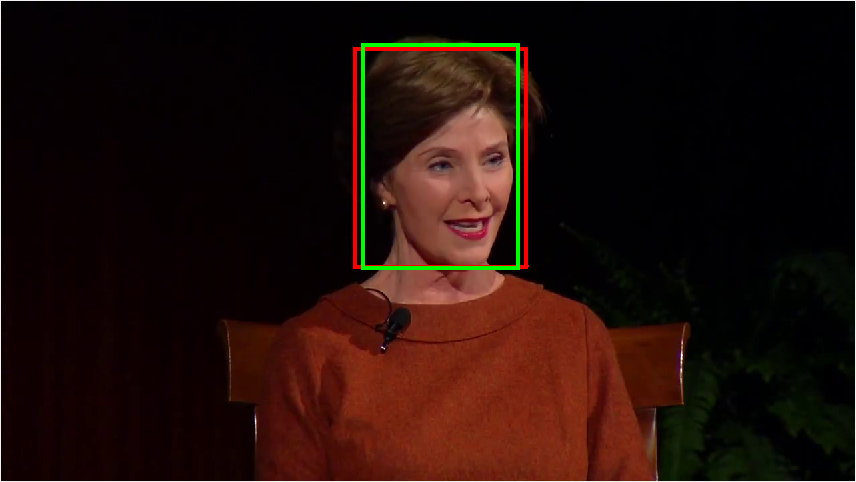} &
   \includegraphics[width=0.16\linewidth,keepaspectratio]{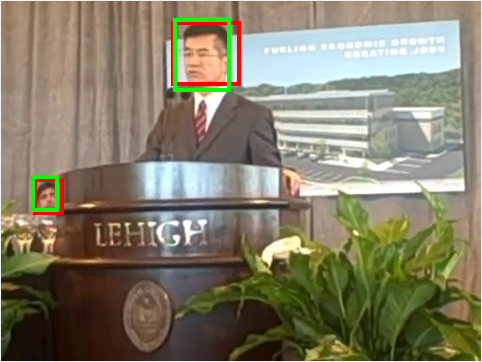} &
   \includegraphics[width=0.16\linewidth,keepaspectratio]{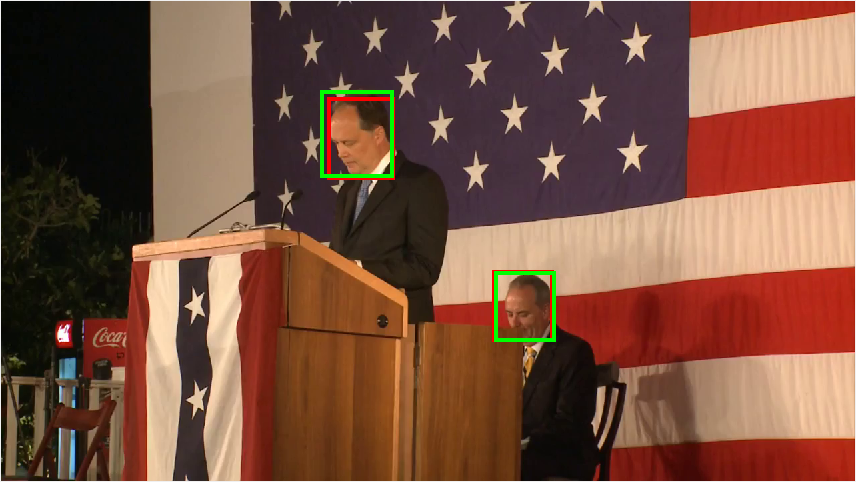} &
   \includegraphics[width=0.16\linewidth,keepaspectratio]{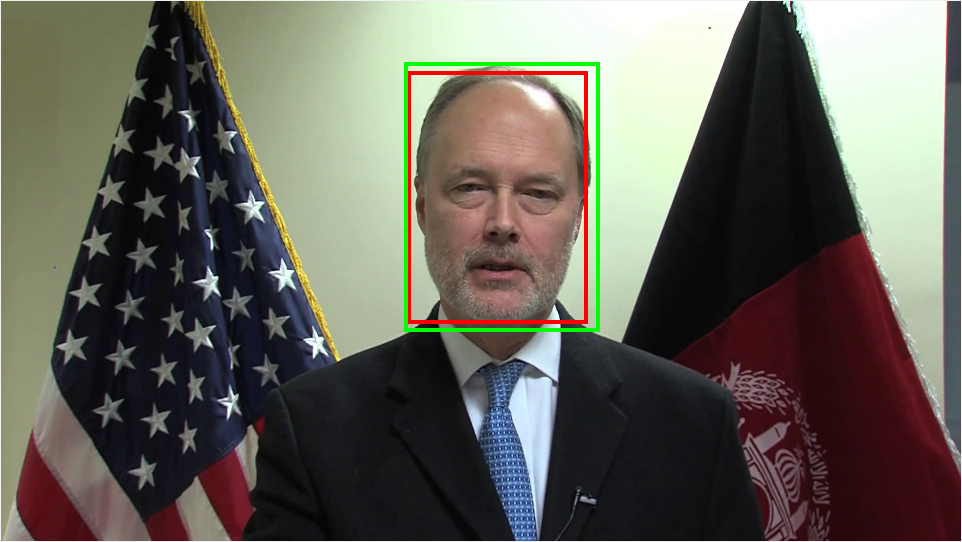} &
   \includegraphics[width=0.16\linewidth,keepaspectratio]{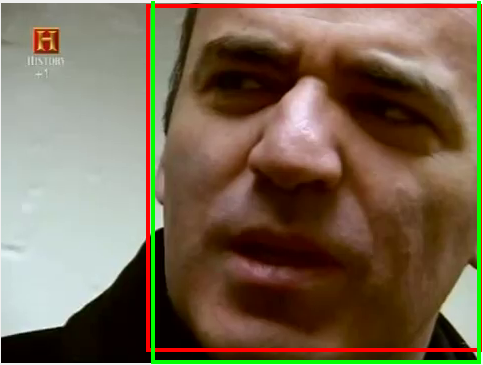} \\
\end{tabular}
\caption{Sample detection results on the IJB-A dataset, where {\color{green}green} bounding boxes are ground-truth annotations and {\color{red} red} bounding boxes are detection results of the Faster R-CNN.}
\label{fig:detsIjba}
\end{figure*}

We follow~\cite{dollar15fast} to measure the detection rate of the top
$N$ proposals by varying the Intersection-over-Union (IoU)
threshold. The larger the threshold is, the fewer the proposals that
are considered to be true objects. Quantitative comparisons of
proposals are displayed in Fig.~\ref{fig:propCompFddb}. As can be
seen, the RPN and DeepBox are significantly better than the other
two. It is perhaps not surprising that learning-based approaches
perform better than the heuristic one, EdgeBox. Although Faceness is
also based on deeply trained convolutional networks (fine tuned from
AlexNet), the rule to compute the faceness score of each proposal is
hand-crafted in contrast to the end-to-end learning of the RPN and
DeepBox. The RPN performs slightly better than DeepBox, perhaps since
it uses a deeper CNN. Due to the sharing of convolutional layers
between the RPN and the Fast R-CNN detector, the process time of the entire
system is lower. Moreover, the RPN does not rely on other object proposal
methods, \eg, EdgeBox.

\subsection{Comparison of Region-based CNN Methods}
We also compare face detection performance of the R-CNN, the Fast
R-CNN, and the Faster R-CNN on FDDB. For both the R-CNN and Fast
R-CNN, we use the top 2000 proposals generated by the Faceness
method~\cite{yang15from}. For the R-CNN, we fine-tune the pre-trained
VGG-M model. Different from the original R-CNN
implementation~\cite{girshick14rich}, we train a CNN with both
classification and regression branches end-to-end
following~\cite{yang15from}. For both the Fast R-CNN and Faster R-CNN, we
fine-tune the pre-trained VGG16 model. As can be observed from
Fig.~\ref{fig:rcnnCompFddb}, the Faster R-CNN significantly outperforms
the other two. Since the Faster R-CNN also contains the Fast R-CNN detector
module, the performance boost mostly comes from the RPN module, which
is based on a deeply trained CNN. Note that the Faster R-CNN also runs
much faster than both the R-CNN and Fast R-CNN, as summarized in
Table~\ref{tab:rcnnComp}.

\subsection{Comparison with State-of-the-art Methods}
Finally, we compare the Faster R-CNN with 11 other top detectors on
FDDB, all published since 2015. ROC curves of the different methods,
obtained from the FDDB results page, are shown in
Fig.~\ref{fig:compSota}. For discrete scores on FDDB, the Faster R-CNN
performs better than all others when there are more than around 200
false positives for the entire test set, as shown in
Fig.~\ref{fig:compSota}(a) and (b). With the more restrictive
continuous scores, the Faster R-CNN is better than most of other
state-of-the-art methods but poorer than
MultiresHPM~\cite{ghiasi14occlusion}. This discrepancy can be
attributed to the fact that the detection results of the Faster R-CNN
are not always exactly around the annotated face regions, as can be
seen in Fig.~\ref{fig:detsFddb}. For 500 false positives, the true
positive rates with discrete and continuous scores are 0.952 and 0.718
respectively. One possible reason for the relatively poor performance
on the continuous scoring might be the difference of face annotations
between WIDER and FDDB.

IJB-A is a relatively new face detection benchmark dataset published
at CVPR 2015, and thus not too many results have been reported on it
yet. We borrow results of other methods
from~\cite{cheney15unconstrained,klare15pushing}. The comparison is
shown in Fig.~\ref{fig:compSota}(d). As we can see, the Faster R-CNN
performs better than all of the others by a large margin.

We further demonstrate qualitative face detection results in
Fig.~\ref{fig:detsFddb} and Fig.~\ref{fig:detsIjba}. It can be observed
that the Faster R-CNN model can deal with challenging cases with
multiple overlapping faces and faces with extreme poses and scales.



\section{Conclusion}
In this report, we have demonstrated state-of-the-art face detection
performance on two benchmark datasets using the Faster R-CNN. Experimental
results suggest that its effectiveness comes from the region proposal
network (RPN) module. Due to the sharing of convolutional layers
between the RPN and Fast R-CNN detector module, it is possible to use
a deep CNN in RPN without extra computational burden.

Although the Faster R-CNN is designed for generic object detection, it
demonstrates impressive face detection performance when retrained on a
suitable face detection training set. It may be possible to further
boost its performance by considering the special patterns of human
faces.

{\small
\bibliographystyle{ieee}
\bibliography{ref}
}

\end{document}